\documentclass[10pt,twocolumn,letterpaper]{article}

\usepackage[pagenumbers]{style} 

\usepackage{graphicx}
\usepackage{amsmath}
\usepackage{amssymb}
\usepackage{booktabs}
\usepackage{diagbox}
\usepackage{makecell}

\usepackage[T1]{fontenc}
\usepackage{mathtools}
\usepackage{multirow}
\usepackage{dblfloatfix}
\usepackage{enumitem}
\usepackage{float}
\usepackage[dvipsnames]{xcolor}
\usepackage{setspace}
\setdisplayskipstretch{0.45}

\usepackage[pagebackref,breaklinks,colorlinks]{hyperref}

\usepackage[capitalize]{cleveref}
\crefname{section}{Sec.}{Secs.}
\Crefname{section}{Section}{Sections}
\Crefname{table}{Table}{Tables}
\crefname{table}{Tab.}{Tabs.}

\def\cvprPaperID{323} 
\def\confName{CVPR}
\def\confYear{2022}

\begin{document}

\title{Causal Representation Learning for Context-Aware Face Transfer}

\author{Gege Gao$^1$ ~~~~~~~~~~~~~ Huaibo Huang$^2$ ~~~~~~~~~~~~~ Chaoyou Fu$^3$ ~~~~~~~~~~~~~ Ran He$^4$ \\
	National Laboratory of Pattern Recognition\\
	Institute of Automation, Chinese Academy of Sciences\\
	{\tt\small $^{1,2}$\{firstname.lastname@cripac.ia.ac.cn\}} ~~~
	{\tt\small $^3$\{chaoyou.fu@nlpr.ia.ac.cn\}} ~~~ 
	{\tt\small $^4$\{rhe@nlpr.ia.ac.cn\}}
	\vspace{-4mm}
}
\maketitle

\begin{abstract}
	Human face synthesis involves transferring knowledge about the identity and identity-dependent shape of a human face to target face images where the context (\eg, facial expressions, head poses, and other background factors) may change dramatically. Human faces are non-rigid, so facial expression leads to deformation of face shape, and head pose also affects the face observed in 2D images.
	A key challenge in face transfer is to match the face with unobserved new contexts, adapting the identity-dependent face shape (IDFS) to different poses and expressions accordingly. 
	In this work, we find a way to provide prior knowledge for generative models to reason about the appropriate appearance of a human face in response to various expressions and poses. We propose a novel context-aware face transfer model, called CarTrans, that incorporates causal effects of contextual factors into face representation, and thus is able to be aware of the uncertainty of new contexts. We estimate the effect of facial expression and head pose in terms of counterfactuals by designing a controlled intervention trial, thus avoiding the need for dense multi-view observations to cover the pose-expression space well. 
	Moreover, we propose a kernel regression-based encoder that eliminates the identity specificity of the target face when encoding contextual information from the target image. The resulting method shows impressive performance, allowing fine-grained control over face shape and appearance under various contextual conditions. 
\end{abstract}

\vspace{-5mm}
\section{Introduction}
\label{sec:intro}
Face transfer aims to transfer knowledge about human face to new 2D face images, including the intrinsic face identity, and appearance properties like face shape that are both identity-dependent and influenced by facial expression and head pose. 
Most existing methods~\cite{Korshunova_2017_ICCV, natsume_2018_rsgan, bao2018towards, Nirkin_2019_ICCV, Li_2020_CVPR, Gao_2021_CVPR} use pre-trained in face models, typically in recognition tasks~\cite{Deng_2019_CVPR}, to represent this knowledge of human faces.

However, face recognition models often constrain the face representations estimated on face images of the same individual under different expressions and poses to be close enough~\cite{Deng_2019_CVPR}, in order to obtain robust performance for accurately identifying human faces in different contexts, 
thus may not encode information about the face appearance properties. 
Therefore, a recognition-oriented face representation does not contain sufficient information for generative modeling the possible IDFS in response to different poses and expressions. 
Also, a pre-trained face representation becomes a deterministic estimate given a specific face image as input, thus is not able to be aware of the new context. 
 
To better match a source face to different target contexts, we propose a novel Context-Aware Representation (CAR) method for human faces in raw 2D images, that explicitly incorporates the causal effect of facial expression and head pose (denoted as, $f^{expo}$) into the face representation as an inductive bias for generative modeling, to reason about the potential changes in face shape in response to different $f^{expo}$, and is therefore able to appropriately adjust the original face representation to match with the target face image. 
The resulting new face representation is no longer deterministic, but takes into account the uncertainty of the target context, and is therefore context-aware.

Moreover, to enable face transfer using unstructured and unposed raw face images, it is important to eliminate the identity-specific information when encoding the contextual representation from the target image. 
To this end, we propose a Kernel-Based Regressive Encoder (KeRE) that runs multiple kernel regressions on the latent features of target images via a set of learnable kernels. 
By modeling a constrained optimization problem, the kernels are trained to perform soft classifications to separate the feature space into the identity-specific information and other contextual information. 
Based on these kernels, KeRE eliminates the identity specificity of the target image by regressive transformations that focus on the identity-dependent subspace, while keeping the contextual information unchanged. 

Combining the context-aware face representation with the kernel regression-based encoder, extensive experiments show that CarTrans can better adapt the source face to characterize its appropriate appearance in various new target contexts. Our contributions are summarized as:

\setlist{nolistsep}
\begin{itemize}[noitemsep]

	\item{We propose CAR, a new context-aware representation method to better generate the appearance of individual faces according to different contexts in target images.}

	\item{We incorporate the causal effect of target $f^{expo}$ into source face representation, and estimate this effect in terms of counterfactuals without requiring any multi-view data to cover the space of expressions and poses.} 

	\item{We propose a kernel-based regressive encoder (KeRE) that eliminates the identity specificity of the target face when representing the target context information.}	

	\item{Experimental results show that incorporating the causal effect of contextual factors as an inductive bias into generative models, paired with kernel-based context encoder, enables fine-grained match with various new contexts across large appearance gaps.}
\end{itemize}

\section{Method} \label{sec:method}

\noindent
\textbf{Symbol.} Denote the source and target face images as $X_s$ and $X_t$, respectively\footnote{
	We use subscript $s$ and $t$ to distinguish between features of $X_s$ and $X_t$, and superscript $(i)$ to sample estimates. $\{\cdot\}$ represents a set of values. A complete {\bf table of notations} can be found in the appendix.
}. 
Face transfer aims at generating a new face image, denoted as $Y_{s,t}$, that shares the face knowledge (\eg, about identity, shape and appearance) with $X_s$ and the context (\eg, facial expression, head pose, and backgrounds) with $X_t$. 
In general, the generation process in previous works can be formulated as: $Y_{s,t} = G(z_s, \{H_t\})$, where $G(\cdot)$ is a neural generator, $z_s$ is the face representation of $X_s$ estimated by a pre-trained face recognition model $M^{id}(\cdot)$~\cite{Deng_2019_CVPR}, \ie, $z_s = M^{id}(X_s)$, and $H_t$ stands for the representation of contextual information in $X_t$. In addition, we use a pre-trained 3D face alignment model, denoted as $M^{3d}(\cdot)$~\cite{guo_2020_3ddfa}, to estimate the facial expression and head pose of the target face, \ie, $f_t^{expo} = M^{3d}(X_t)$. 

Our model, CarTrans, makes improvements to both $z_s$ and $F_t$. 
The Context-Aware Representation (CAR) for the source face to adapt to the target context is formulated as: 
\begin{equation}
	z^*_{s,t} = \lambda_{\Theta}(z_s, \Delta z_t) , 
	\label{eq:car}
\end{equation}
where $z^*_{s,t}$ is the resulting context-aware representation for the source face, $\Delta z_t$ is a causal corrective of $z_s$ according to the target context, and $\lambda_{\Theta}$ stands for a neural function. 

In this section, first a well-designed intervention trial is introduced to give an estimation of $\Delta z_t$, and thus obtain $z^*_{s,t}$. 
Then, we explain how the Kernel-based Regressive Encoder (KeRE) extracts contextual information from raw target images while eliminating the identity specificity of target faces. On this basis, we combine the identity-agnostic target contextual representation with the context-aware source face representation to produce the final result $Y_{s,t}$. 
For simplicity, we remove the subscripts $s$ and $t$ when there is no difference between the processing of the source and target features in the following. 
We begin with an analysis on the causes of $\Delta z_t$ in target context.


\subsection{Context-Aware Face Representation}
\subsubsection{Causal Analysis on $\Delta z_t$}
\label{subsec:causal_analysis}
Recognition-oriented face representations are often trained not to encode information about face shape, as this information is easily affected by $f^{expo}$. 
Therefore, in the context of face synthesis, additional information is needed in addition to those encoded by $M^{id}(\cdot)$ (with fixed parameters) for correctly adapting IDFS of an individual to various $f^{expo}$ in new contexts. 
We infer such additional information causally from $f^{expo}$ of the target context.

\textbf{Causal assumption.} We assume that a deterministic face representation $z$ is influenced by facial expression and head pose $f^{expo}$ (in 2D domain) in the context of face synthesis problems. Also, there are certain unobserved factors like facial occlusions or background noise, denoted as $u^{bg}$. The graphical formalism of causal assumption is shown in Fig.~\ref{fig:dag-a}. The formal expression of this assumption can be found in the appendix. 
Based on this, we model the corrective $\Delta z_t$ as the causal effect of $f^{expo}$ and $u^{bg}$ on $z$.  
A revisit of causal inference can be found in the appendix. 

\begin{figure}[t]  
	\centering
	\vspace{-2mm}
	\begin{subfigure}{0.4\linewidth}
		\centering
		\includegraphics[width=0.68\linewidth]{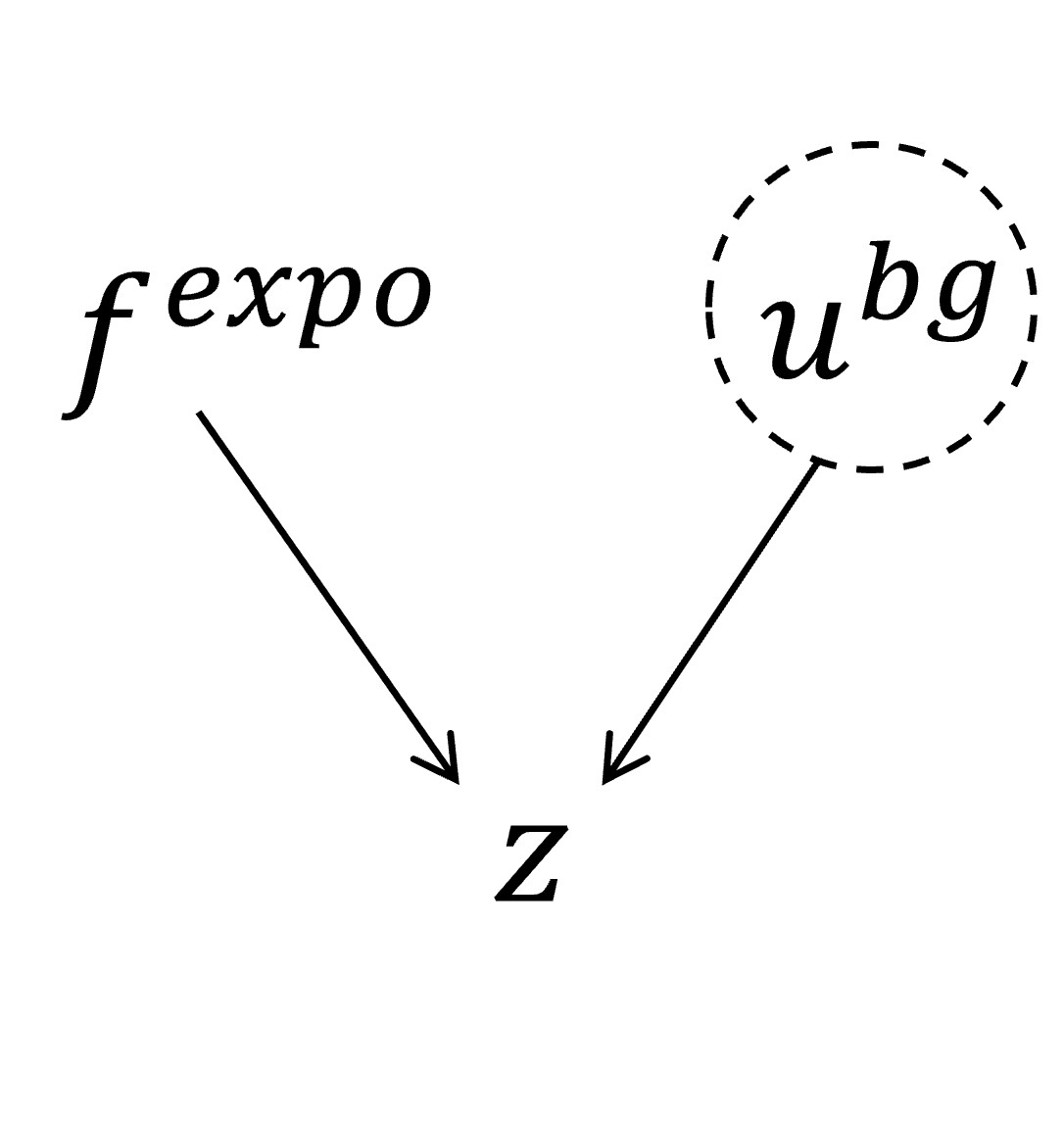}
		\caption{Initial DAG. $u^{bg}$: unobserved causal factors. $A$$\rightarrow$$B$ means that $A$ has a causal effect on $B$. }
		\label{fig:dag-a}
	\end{subfigure}
	\hfill
	\begin{subfigure}{0.55\linewidth}
		\includegraphics[width=1.0\linewidth]{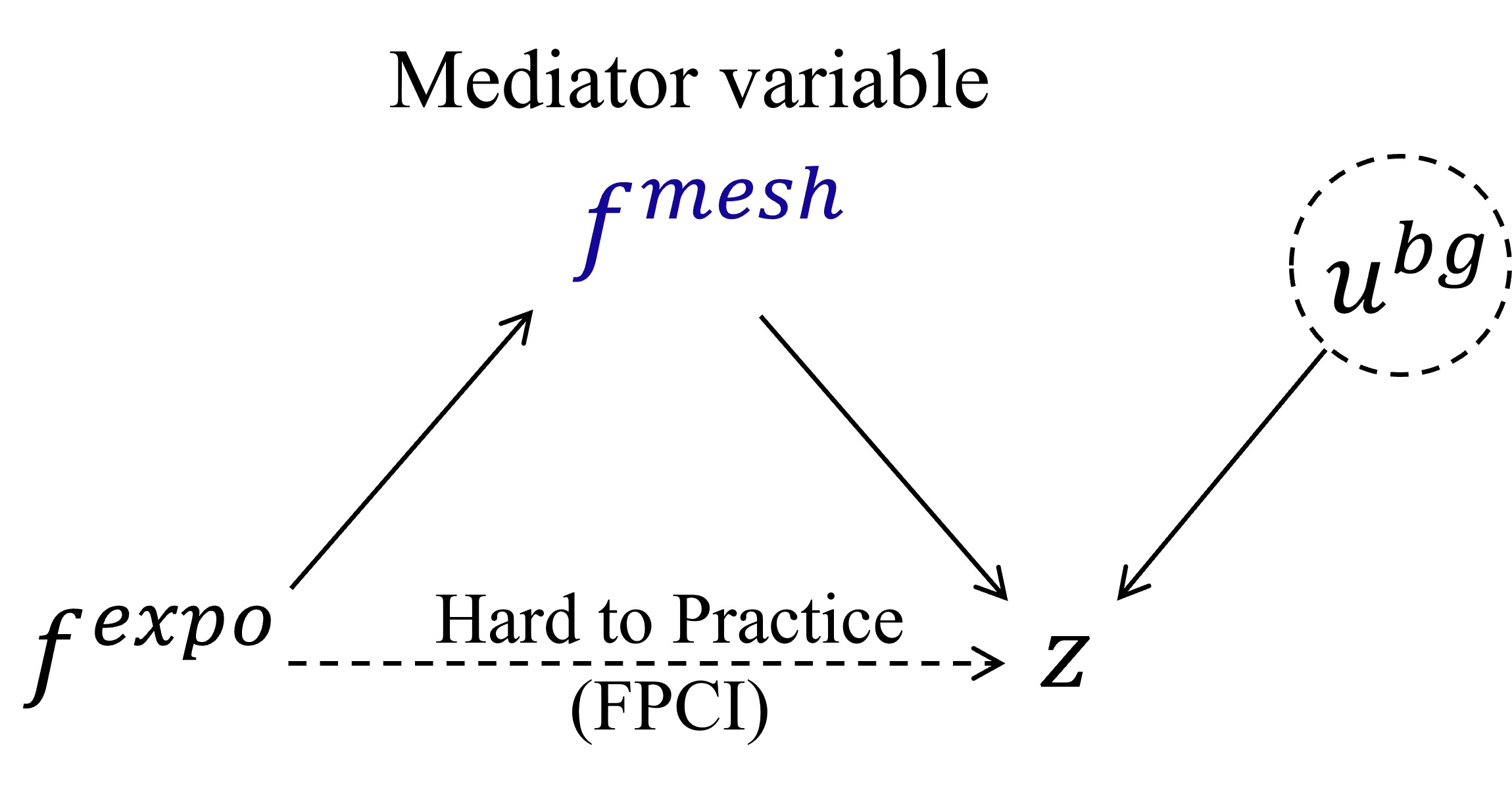}
		\caption{Final DAG of causal assumption. We mediate in the initial causality $f^{expo} \rightarrow z$, arriving at two new causal links. $f^{mesh}$: 2D dense face meshes, used as a measurement of IDFS.}
		\label{fig:dag-b}
	\end{subfigure}
	\vspace{-1mm}
	\caption{
		\textbf{Causal assumption.} 
		The FPCI (fundamental problem of causal inference~\cite{holland_1986_statistics}) in (b) refers to a common dilemma where the controlled trail is infeasible. We overcome this problem by introducing $f^{mesh}$ as a mediator variable. 
		Note that the variables $\{z, f^{expo}, f^{mesh}\}$ in both graphs are all from same individuals. }
	\label{fig:dag}
	\vspace{-3mm}
\end{figure}

\textbf{Controlled intervention trial. } 
In general, it is difficult to directly estimate causal effects from observational data~\cite{rubin_1974_estimating}. Therefore, we design a controlled intervention trial to estimate the causal effect of $f^{expo}$ in terms of counterfactual:  
\textbf{(i) Controlled group (CG)} are the original outcomes of $z = M^{id}(X)$ directly computed from raw face images. \textbf{(ii) Treatment group (TG)} are the potential outcomes of $z$ unaffected by information about $f^{expo}$. 
Then, based on the Rubin Causal Model (RCM)~\cite{rubin_1974_estimating, rubin_2005_causal}, the causal effect of $f^{expo}$ on $z$ is measured by the difference in the outcomes between the two groups. 

\textbf{Inference via mediator.} In practice, we slightly modify the initial causal model. 
As shown in Fig.~\ref{fig:dag-b}, we use IDFS (identity-dependent face shape), measured by 2D dense face meshes $f^{mesh}$, as a \emph{mediator variable}~\cite{pearl_2018_book}, since it is easier to measure the effect of $f^{expo}$ on the face shape than on the highly-abstracted face representation $z$ (\eg, a high dimensional embedding). 
Intuitively, the changes in $\{z^{(i)}\}$ of the same individual can also be attributed to the non-rigid deformation of the observed face shape in 2D images, caused by facial motions $f^{expo}$. 
We use $f^{mesh}$ to evaluate such deformation of face shape, so $f^{mesh}$ also has an effect on $z$ estimates, with $f^{expo}$ confounding both $z$ and $f^{mesh}$. 
Therefore, we introduce $f^{mesh}$ as the mediator and converts the causality $f^{expo} \rightarrow z$ into \textbf{two new causal links}: 
\\
$\bullet$ $f^{expo} \rightarrow f^{mesh}$: From the technical perspective, the key step lies in the first causal link, that is to estimate the counterfactual outcome of $f^{mesh}$ unaffected by information about $f^{expo}$ (for TG), as described in Section~\ref{subsec:counter}.  
\\
$\bullet$ $f^{mesh} \rightarrow z$:  
we translate the effect of $f^{expo}$ from the 2D face shape $f^{mesh}$ to the final face representation $z$ using a parameterize neural function, and the effect of $u_{bg}$ is estimated by simulation, as described in Section~\ref{subsec:translate}.  

\begin{figure}[t]  
	\centering
	\vspace{-2mm}
	\includegraphics[width=1.0\linewidth]{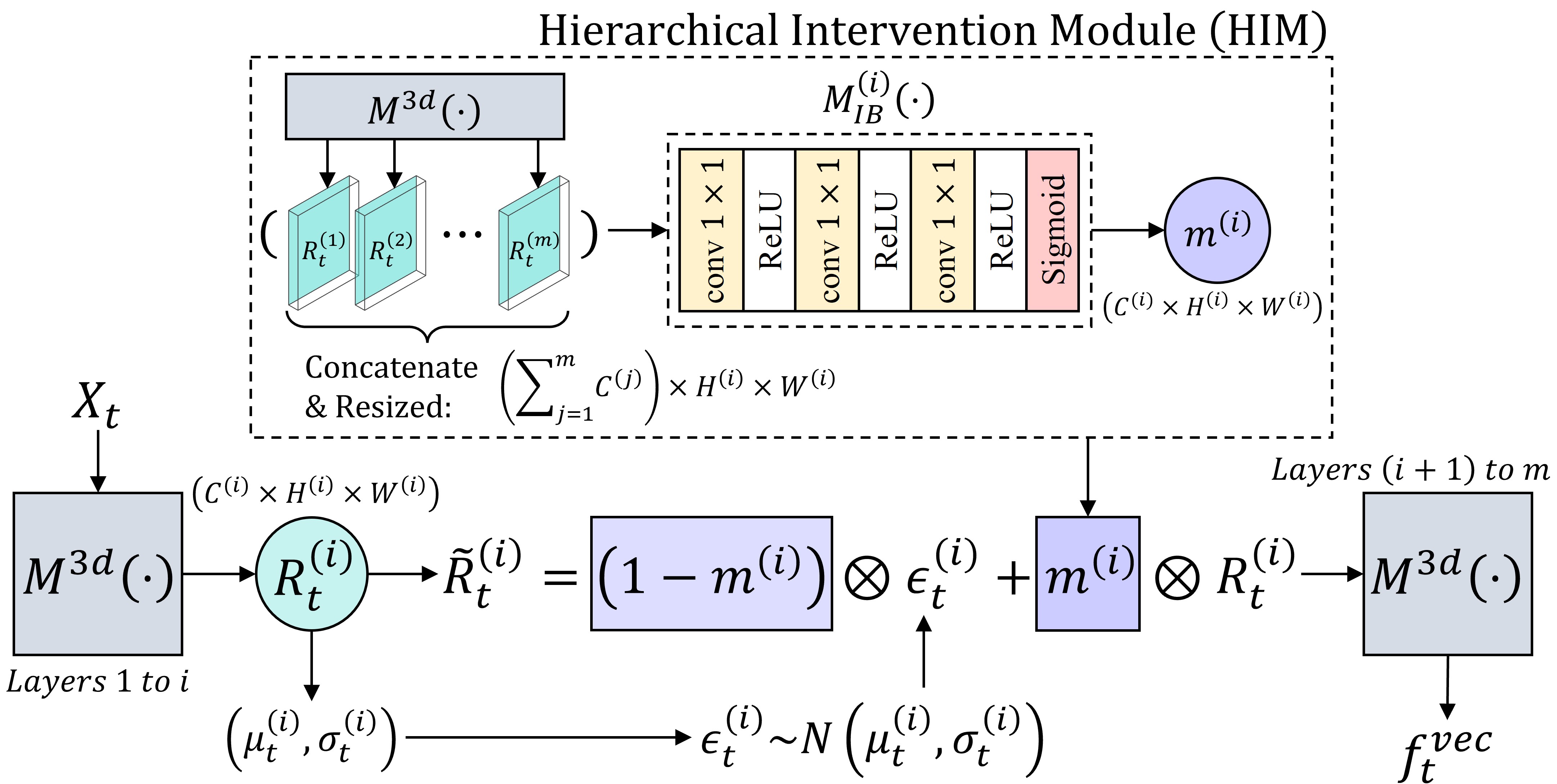}
	\caption{\textbf{HIM.} The Hierarchical Intervention Module (HIM) is designed for constructing counterfactual face meshes that are maximally unaffected by $f^{expo}$. For details please see Section~\ref{subsec:counter}.}
	\vspace{-3mm}
	\label{fig:net-a}
\end{figure}

\vspace{-2mm}
\subsubsection{Counterfactual Inference: $f^{expo} \rightarrow f^{mesh}$}
\label{subsec:counter}
Given a target face image $X_t$, we first use the pre-trained $M^{3d}(\cdot)$ to estimate face meshes, obtaining the original outcome $f_t^{mesh}$ for the controlled group (CG). 
To model the counterfactual face meshes that are maximally unaffected by $f_t^{expo}$ for the treatment group (TG), we design a Hierarchical Intervention Module (HIM) based on the information bottle principle~\cite{Tishby_2015_IB, Ravid_2017_opening, Schulz_2020_Restricting} in a novel hierarchical fashion. 
We introduce external interventions on $f_t^{expo}$ to filter out information about $f_t^{expo}$ when computing $f_t^{mesh}$. A revisit of information bottle principles can be found in the appendix.  

Concretely, as shown in Fig.~\ref{fig:net-a}, after feeding forward $X_t$, we extract a set of $m$ intermediate representations from $M^{3d}(\cdot)$, denoted as $\{R_t^{(i)}\}$ ($i = 1, \cdots, m$). Then, for each $R_t^{(i)}$, we build an information bottleneck trade-off in the latent space of $R_t^{(i)}$, successively from $i=1$ to $i=m$, based on a \textbf{filtering mask} $m^{(i)}$. 
Each mask is predicted by a tiny convolutional block $M_{IB}^{(i)}(\cdot)$ individually. To give $M_{IB}^{(i)}(\cdot)$ a global view over the computing process of $M^{3d}(\cdot)$ in predicting $m^{(i)}$, we use all $\{R_t^{(i)}\}$ as input: 
\begin{equation}
	m^{(i)} = M_{IB}^{(i)}(R_t^{(1)}, R_t^{(2)}, \cdots, R_t^{(m)}),  
	\label{eq:mask}
\end{equation} 
where $\{R_t^{(1)}, \cdots, R_t^{(m)}\}$ are resized spacial-wise according to $R_t^{(i)}$ and concatenated together channel-wise, in a shape of $(\sum_{j=1}^{m}C^{(j)}) \times H^{(i)} \times W^{(i)}$ before being fed into $M_{IB}^{(i)}(\cdot)$. 
The resulting $m^{(i)}$ is of the same shape as $R_t^{(i)}$, with values in the range of $0 \sim 1$ (after \emph{sigmoid} activation). 
Then, based on $m^{(i)}$, we inject additive Gaussian noise $\epsilon_t^{(i)}$ into each latent space of $R_t^{(i)}$ to filter out part of information $R_t^{(i)}$ contains, leading to a \textbf{compressed version} as $\widetilde{R}^{(i)}_t$\footnote{
	We use a tilde term $\widetilde{V}$ to represent a compressed version of variable $V$ with some of the information in $V$ been replaced by white noises.
}: 
\begin{equation}
	\widetilde{R}^{(i)}_t = m^{(i)} \otimes R_t^{(i)} + (1 - m^{(i)}) \otimes \epsilon^{(i)}_t,  
	\label{eq:compress}
\end{equation}
where $\otimes$ denotes element-wise multiplication. $\epsilon_t^{(i)}$ is of the same size as $R_t^{(i)}$, sampled from a Gaussian empirical distribution based on the mean and variance of $R_t^{(i)}$, \ie, $\epsilon_t^{(i)} \sim \mathcal{N}(\mu_t^{(i)}, \sigma_t^{(i)})$, since activations after convolutional layers tend to have a Gaussian distribution~\cite{klambauer_2017_self}. 

The mask $m^{(i)}$ (with values between $0$ and $1$) is designed to show the importance of corresponding neurons in each $R_t^{(i)}$ in terms of predicting $f_t^{expo}$, \ie, how \textit{informative} they are for $f_t^{expo}$. 
Initially, $\{M_{IB}^{(i)}(\cdot)\}$ are not yet learned to properly predict such masks, and thus noises are randomly injected into $\{R_t^{(i)}\}$ at first. 
Therefore, the key insight lies in how to supervise the training process of each $M_{IB}^{(i)}(\cdot)$. 
To this end, we incorporate a set of trade-offs in the latent spaces of $\{R_t^{(i)}\}$ into the training objective as:  
\begin{equation}
	\mathcal{L}_{mask} = MI(\widetilde{R}^{(i)}_t; R_t^{(i)}) + \alpha \cdot \mathcal{L}_{expo},  
	\label{eq:loss_mask}
\end{equation}
where $\alpha$ ($>0$) is a hyperparameter. $\mathcal{L}_{mask}$ consists of two loss terms that play against each other: 
\\
$\bullet$ $MI(\widetilde{R}^{(i)}_t; R_t^{(i)})$ is defined as the mean of mutual information between $\{R_t^{(i)}\}$ and their compressed version $\{\widetilde{R}^{(i)}_t\}$:  
\begin{equation}
	MI(\widetilde{R}^{(i)}_t; R_t^{(i)}) = \frac{1}{m} \sum_{i=1}^{m} I(\widetilde{R}^{(i)}_t; R_t^{(i)}), 
\end{equation}
where $I(\cdot ; \cdot)$ stands for the mutual information. Therefore, $MI(\widetilde{R}^{(i)}_t; R_t^{(i)})$ measures how much information in $\{R_t^{(i)}\}$ is replace by noises, thus minimizing $\mathcal{L}_{mask}$ tends to \textbf{maximally compress} the information contained in $\{R_t^{(i)}\}$. While at the same time, 
\\
$\bullet$ $\mathcal{L}_{expo}$ is designed to \textbf{maximally preserve} the information in $\{R_t^{(i)}\}$ that is important for predicting $f_t^{expo}$. We use it to measure the difference in $f_t^{expo}$ before and after injecting the noises to $\{R_t^{(i)}\}$. 
Concretely, each time we replace one $R_t^{(i)}$ with its compressed version $\widetilde{R}^{(i)}_t$, and recompute later ones $\{R_t^{(k)}\}$ ($k>i$) based on $\widetilde{R}^{(i)}_t$, from $i=1$ to $i=m$ hierarchically. In this way, each $m^{(i)}$ works on the basis of the previous ones, so as to introduce the interventions continuously. 
Denote the output after $R_t^{(m)}$ being replaced $\widetilde{R}^{(m)}_t$ as $\widetilde{f}_t^{expo}$, then $\mathcal{L}_{expo}$ is defined as:
\begin{equation}
	\mathcal{L}_{expo} = \|f_t^{expo} - \widetilde{f}_t^{expo} \|_1.
\end{equation} 

By minimizing $\mathcal{L}_{mask}$, $MI(\widetilde{R}^{(i)}_t; R_t^{(i)})$ tends to force $m^{(i)}$ close to $0$, while in contrast, $\mathcal{L}_{expo}$ tends to force $m^{(i)}$ close to $1$. 
In this way, values in $m^{(i)}$ corresponding to neurons informative of $f^{expo}$ are brought near $1$, while others are brought near $1$.  
Thus, by hierarchically replacing each $R_t^{(i)}$ in $M^{3d}(\cdot)$ with another compressed version of $R_t^{(i)}$:
\begin{equation}
	E^{(i)}_t = m^{(i)} \otimes \epsilon^{(i)}_t + (1 - m^{(i)}) \otimes R_t^{(i)},  
	\label{eq:n-compress}
\end{equation}
which is defined just in contrast to $\widetilde{R}^{(i)}_t$ in Eq.~\ref{eq:compress}, we will arrive at the \textbf{counterfactual face meshes} that are maximally unaffected by information about $f_t^{expo}$ for TG, and thus enabling to estimate the effect of $f_t^{expo}$ according to RCM. 

\vspace{-3mm}
\subsubsection{Causal Effect Translation: $f^{mesh} \rightarrow z$}
\label{subsec:translate}

\begin{figure}[t]  
	\centering
	\vspace{-2mm}
	\includegraphics[width=0.95\linewidth]{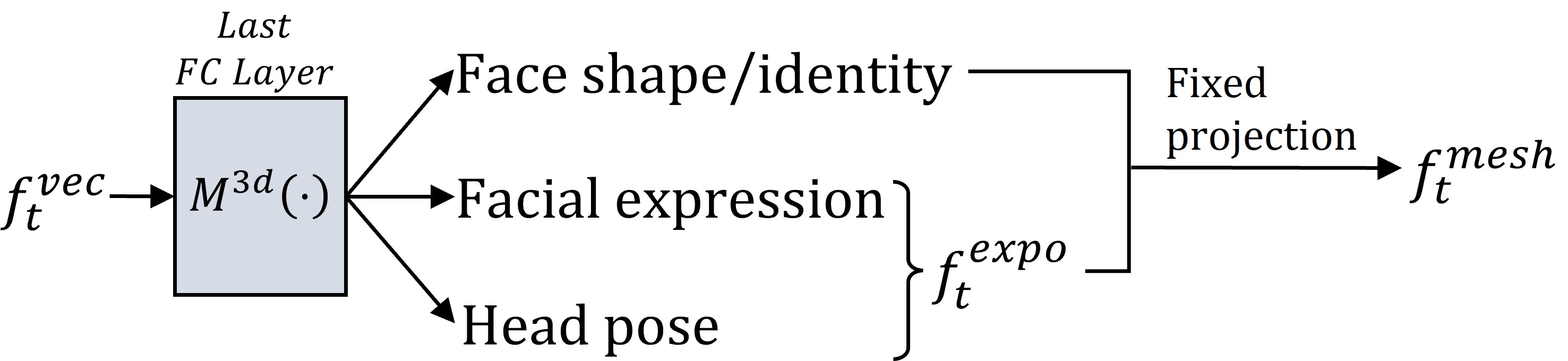}
	\caption{Face meshes $f_t^{mesh}$ are mapped from the latent vector $f_t^{vec}$ before the last FC layer in $M^{3d}(\cdot)$. Both the projection (via a pre-trained BFM~\cite{bfm_2018}) and the FC layer are fixed (Section~\ref{subsec:translate}).  
	}
	\vspace{-3mm}
	\label{fig:net-b}
\end{figure}

\textbf{Effect estimation.} 
To make the entire process of CarTrans differentiable to enable end-to-end training,
in practice, we estimate the causal effect of $f_t^{expo}$ based on potential outcomes of the latent vector $f_t^{vec}$ (which is $1024$-dimensional) before the last fully-connected layer in $M^{3d}(\cdot)$, instead of the discrete face meshes, since the mapping from $f_t^{vec}$ to $f_t^{mesh}$ is fixed, as shown in Fig.~\ref{fig:net-b}. 
Denote the counterfactual outcome of the latent vector after hierarchically replacing $\{R_t^{(i)}\}$ with $\{E^{(i)}_t\}$ (in Eq.~\ref{eq:n-compress}) as $\widetilde{f}_t^{vec}$. According to RCM, the causal effect of $f_t^{expo}$ is formulated as:
\begin{equation}
	\delta(f^{expo} \rightarrow f^{mesh}) = \widetilde{f}_t^{vec} - f_t^{vec}. 
	\label{eq:link-a}
\end{equation}
\textbf{Effect translation.} We then translate the causal effect $\delta(f^{expo} \rightarrow f^{mesh})$ from IDFS to $z$, arriving at the final estimation of the causal corrective $\Delta z_t$, formulated as: 
\begin{equation}
	\Delta z_t = g_{\theta}(\widetilde{f}_t^{vec} - f_t^{vec}) + u_t^{bg},  
	\label{eq:delta_z}
\end{equation}
where $g_{\theta}(\cdot)$ with parameters $\theta$ is a neural function. In practice, we set $g_{\theta}(\cdot)$ as a fully-connected layer that maps $\widetilde{f}_t^{vec} - f_t^{vec}$, a $1024$-dimensional vector, to the $512$-dimensional vector space of $z$. 
The unobserved causal factor $u_t^{bg}$ is simulated by random sampling from the von Mises-Fisher (vMF) distribution (with parameters pre-estimated on a small subset of the training set), since normalized face embedding vectors are often considered to have such a distribution type. We thus simulate $u_t^{bg}$ following the method of~\cite{davidson_2018_hyperspherical, xu_2018_spherical}. 
More details on simulating $u_t^{bg}$ can be found in the appendix.

\begin{figure}[t]
	\vspace{-2mm}
	\centering
	\includegraphics[width=0.85\linewidth]{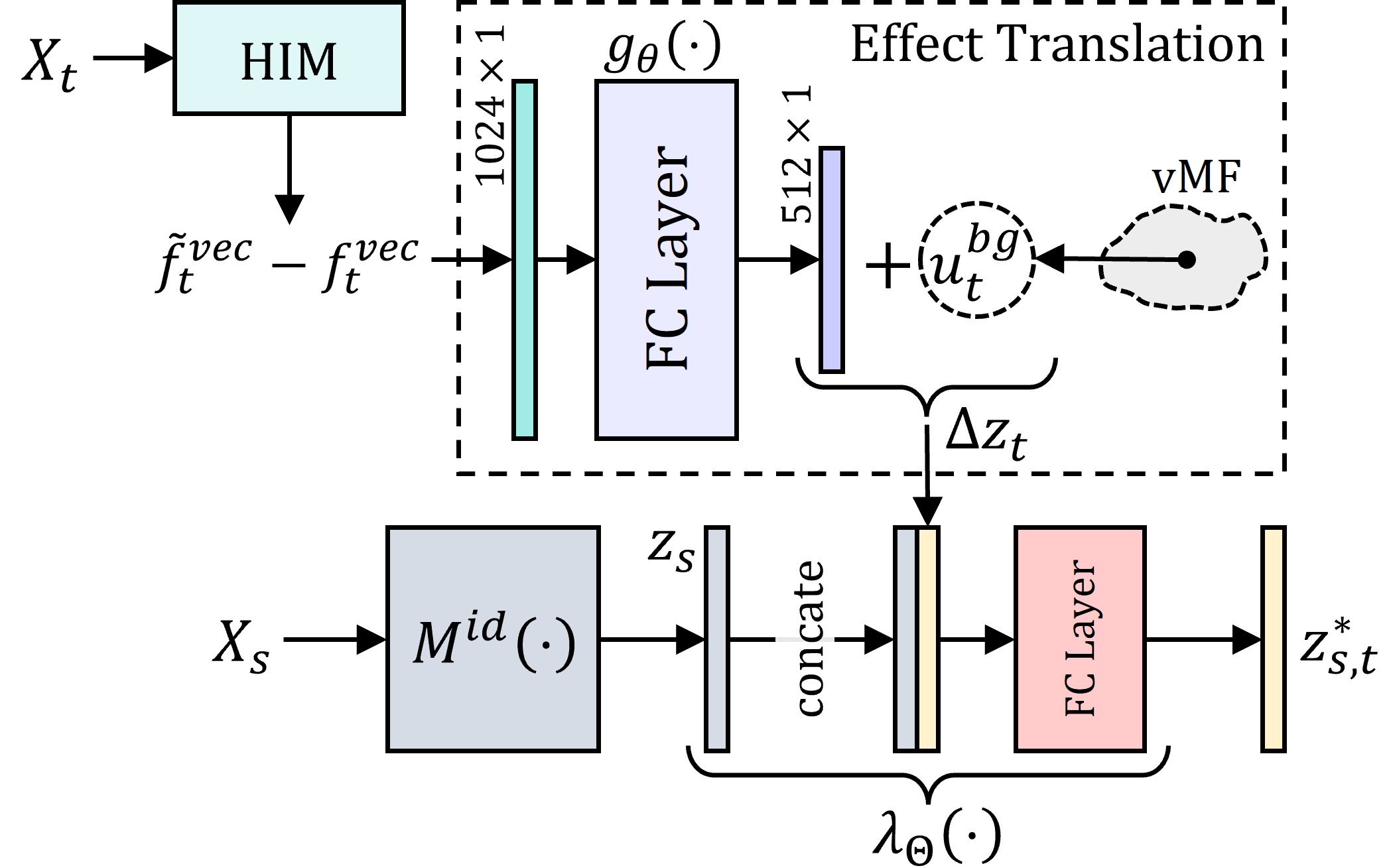}
	\vspace{-2mm}
	\caption{\textbf{CAR.} A pipeline for modeling the context-aware face representation of $X_s$. The new representation $z^*_{s,t}$ takes into account possible contextual factors in target image $X_t$, by incorporating the causal effect of target context, \ie, $\Delta z_t$, with original representation $z_s$ via a neural function $\lambda_{\Theta}$.  
	}
	\vspace{-3mm}
	\label{fig:car}
\end{figure}

Finally, we arrive at the new representation of source face $z^*_{s,t}$, which takes into account the effect of new context in the target image $X_t$, by combining $z_s$ with $\Delta z_t$. A high-level overview of modeling $z^*_{s,t}$ is shown in Fig.~\ref{fig:car}.

\subsection{Kernel-based Regressive Encoder}
\label{subsec:KeRE}

This section describes how the Kernel-based Regressive Encoder (KeRE) learns contextual representations from raw target image $X_t$ while eliminating the identity specificity of the target face, by exploring latent spaces of the $M^{id}(\cdot)$, \ie, a pre-trained ResNet-50 network~\cite{Deng_2019_CVPR} in practice.

The KeRE contains several \textbf{learnable kernels} which are used to perform soft classifications in the spaces of $M^{id}(\cdot)$. 
Given $X_t$ as input, we denote the internal features (after each \textit{ResBlock} from low to mid-level depth) in $M^{id}(\cdot)$ as $\{F_t^{(i)}\}$ ($i=1, 2, \cdots, k$). 
Each $F_t^{(i)}$ contains both the contextual information (\eg, about $f^{expo}$, lighting, and other backgrounds), and the identity-specific information, since the latent space of $F_t^{(i)}$ is less compressed and abstractive~\cite{gafni_2019_live} compared to the final output $z_t = M^{id}(X_t)$. 

Concretely, denote these learnable kernels as $\{k^{(i)}\}$ ($i=1, 2, \cdots, k$). We then build a set of $k$ regressing functions, denoted as $\{r^{(i)}(\cdot)\}$, which act on the internal features $\{F_t^{(i)}\}$ via these kernels:  
\begin{subequations}\label{eq:H}
	\begin{alignat}{2}
		&H_t^{(i)} = r^{(i)}(k^{(i)}\otimes F_t^{(i)}) + (1 - k^{(i)}) \otimes F_t^{(i)}, \label{eq:feature_H}
		\\
		&{\overline{H}}_t^{(i)} = r^{(i)}((1 - k^{(i)})\otimes F_t^{(i)}) + k^{(i)} \otimes F_t^{(i)}, \label{eq:feature_nH}
	\end{alignat}
\end{subequations}
where each $r^{(i)}(\cdot)$ consists of $3$ convolutional layers, $k^{(i)}$ is of the same size as $F_t^{(i)}$ and takes values between $0$ and $1$. 

Our goal is to use the kernel $k^{(i)}$ 
to identify identity-specific information from the latent space of $F_t^{(i)}$ (with its values standing for the probabilities), \ie, 
use $k^{(i)}\otimes F_t^{(i)}$ to represent the identity-specific information, 
and $(1 - k^{(i)}) \otimes F_t^{(i)}$ the contextual information in latent space of $F_t^{(i)}$. 
So that based on $k^{(i)}$, we can use $r^{(i)}(\cdot)$ to smooth away the identity-specific neurons in $F_t^{(i)}$, without changing other neurons that are mainly related to contextual information. 

To this end, we build a constrained optimization model:
\begin{subequations}\label{eq:cop}
	\begin{alignat}{2}
		&\min_{k^{(i)},r^{(i)}} 1 - \cos\langle \hat{z}, z(H_t^{(i)})\rangle, \label{sub-eq-1:1}
		\\
		&\text{subject to: } \cos\langle z_t, z({\overline{H}}_t^{(i)})\rangle = 1. \label{sub-eq-1:2}
	\end{alignat}
\end{subequations}
where $z(V) = M^{id}(X\mid F^{(i)} := V)$, standing for the estimate of a face image $X$, with the $i$-th internal feature $F^{(i)}$ in $M^{id}(\cdot)$ been replaced with an \textit{alternative feature} $V$, $\cos\langle\cdot, \cdot\rangle$ is the cosine similarity between two vectors, 
and $\hat{z}$ is the empirical mean of $z$ estimated on a subset of training data, standing for the representation of an "average face".   

\textbf{On the one hand}, we want to smooth out the identity specificity of $X_t$, so the objective Eq.~\ref{sub-eq-1:1} aims to bring the resulting estimate $z(H_t^{(i)})$, based on the alternative feature $H_{(i)}^t$, close to the average face representation $\hat{z}$ in terms of its cosine similarity. 
\textbf{On the other hand}, we need to keep the context information in $X_t$, so the constraint Eq.~\ref{sub-eq-1:2} requires that the other estimate $z({\overline{H}}_t^{(i)})$, which is instead based on ${\overline{H}}_t^{(i)}$, remains as $z_t$. 

Initially, values in $k^{(i)}$ are all set to $0$. According to Eq.~\ref{eq:feature_nH}, the regression $r^{(i)}(\cdot)$ works on the entire $F_t^{(i)}$, leading to a large change in the output $z({\overline{H}}_t^{(i)})$ compared to $z_t$, \ie, $\cos\langle z_t, z({\overline{H}}_t^{(i)})\rangle$ is significantly different from $1$. 
Also, according to Eq.~\ref{eq:feature_H}, $z(H_t^{(i)})$ is initially equal to $z_t$, which differs significantly from $\hat{z}$. 
Rewrite the constrained optimization model (Eq.~\ref{eq:cop}) in the form of loss function as: 
\begin{equation}
	\begin{aligned}
		\mathcal{L}_{kernel} = \frac{1}{k} \sum_{i=1}^{k} \{2 &- \cos\langle \hat{z}, z(H_t^{(i)})\rangle \\ 
		&- \cos\langle z_t, z({\overline{H}}_t^{(i)})\rangle\}.
	\end{aligned} 
	\label{eq:loss_kernel}
\end{equation}
Then, by training with Eq.~\ref{eq:loss_kernel}, we can effectively supervise the learning process of these kernels $\{k^{(i)}\}$ to correctly identify identity-specific information in the latent spaces of $M^{(id)}(\cdot)$, arriving at a set of $k$ identity-agnostic contextual representations of the target face image as $\{H_t^{(i)}\}$. 

\subsection{Generation for Final Result}

\begin{figure}[t]
	\centering
	\vspace{-3mm}
	\includegraphics[width=0.99\linewidth]{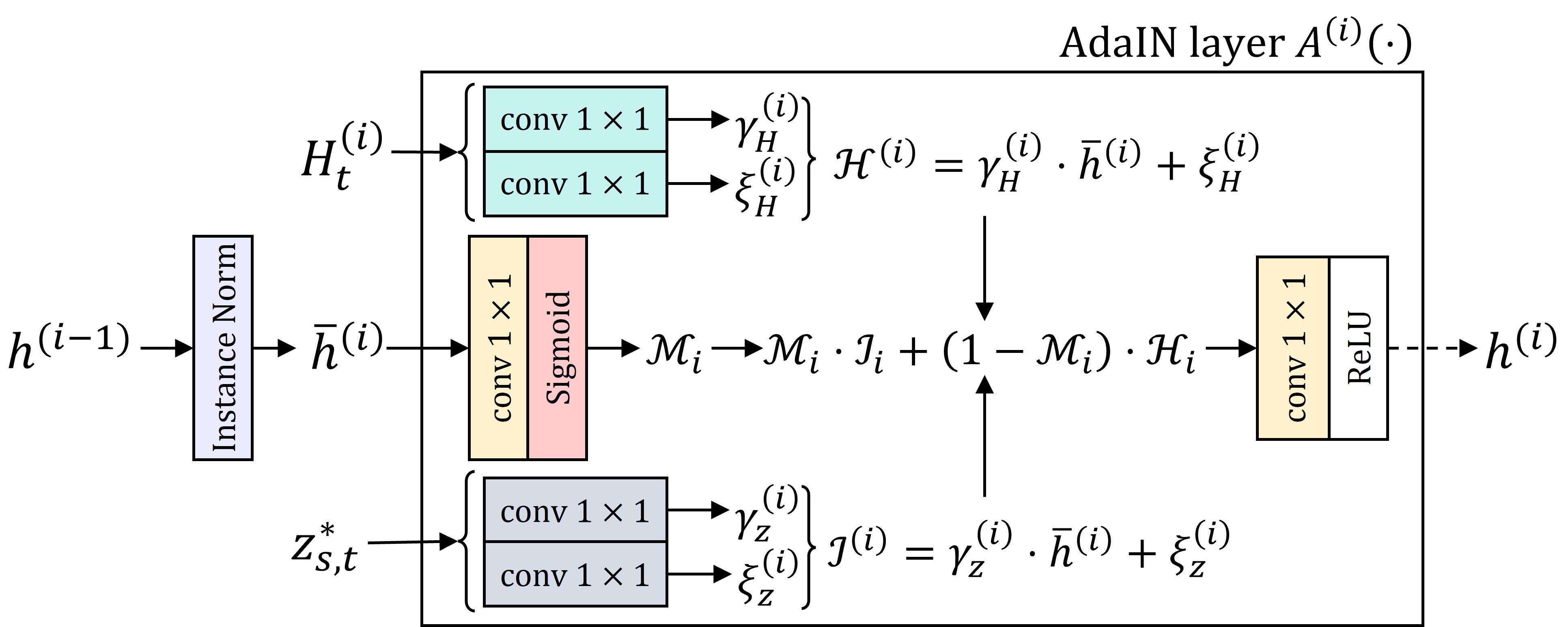}
	\caption{\textbf{AdaIN layer.} Our generator $G(\cdot)$ consists of $k$ AdaIN layers. Starting from $h^{(0)} = z^*_{s,t}$, each AdaIN layer $A^{(i)}(\cdot)$ works successively based on the output of $A^{(i-1)}(\cdot)$, taking $z^*_{s,t}$ and $H_t^{(i)}$ as input ($i=1,2,\cdots,k$). 
	}
	\vspace{-3mm}
	\label{fig:adain}
\end{figure}

Finally, the resulting face image $Y_{s,t}$ is generated by combining the context-aware face representation $z^*_{s,t}$ with the identity-agnostic contextual representations $\{H_t^{(i)}\}$:
\begin{equation}
	Y_{s,t} = G(z^*_{s,t}, \{H_t^{(i)}\}), 
	\label{eq:generation}
\end{equation}
where the generator $G(\cdot)$ contains a set of $k$ Adaptive Instance Normalization (AdaIN) layers~\cite{Huang_2017_ICCV, Park_2019_CVPR, Li_2020_CVPR, Gao_2021_CVPR} that work successively. Each AdaIN layer, denoted as $A^{(i)}(\cdot)$ ($i=1,2,\cdots,k$), takes $z^*_{s,t}$ and $H_t^{(i)}$ as input, working based on the output of its previous layer: 
\begin{equation}
	h^{(i)} = A^{(i)}(z^*_{s,t}, H_t^{(i)} \mid h^{(i-1)}), 
\end{equation}
where $h^{(0)} = z^*_{s,t}$, as shown in Fig.~\ref{fig:adain}. $Y_{s,t}$ is obtained by up-sampling the output after the last AdaIN layer $A^{(k)}(\cdot)$. 

\subsection{End-to-End Training}

In addition to $\mathcal{L}_{mask}$ and $\mathcal{L}_{kernel}$ defined in Eq.~\ref{eq:loss_mask} and Eq.~\ref{eq:loss_kernel}, 
we further supervise the identity information of the generated face in $Y_{s,t}$ via an \textbf{identity loss} formulated as:
\begin{equation}
	\mathcal{L}_{id} = 1 - \cos\langle M^{id}(Y_{s,t}), sg[z^*_{s,t}]\rangle,
	\label{loss:id}
\end{equation}
where $sg[\cdot]$ stands for the stop-gradient operator~\cite{Oord_2017_vqvae}.

Also, we extract internal features $\{F_{Y_{s,t}}^{(i)}\}$ out when computing $M^{id}(Y_{s,t})$ to supervise the contextual information in $Y_{s,t}$, by designing a \textbf{context loss} based on the kernels $\{k^{(i)}\}$ (described in Section~\ref{subsec:KeRE}) as: 
\begin{equation}
	\mathcal{L}_{context} = \frac{1}{k} \sum_{i=1}^{k} [(1 - sg[k^{(i)}]) \otimes (F_t^{(i)} - F_{Y_{s,t}}^{(i)})].
	\label{loss:context} 
\end{equation}

Moreover, we use a multi-scale discriminator~\cite{karras_2020_analyzing}, denoted as $D(\cdot)$, along with an \textbf{adversarial loss} $\mathcal{L}_{adv}$~\cite{Alexia_2018_RSGAN}, to make the generated $Y_{s,t}$ more realistic: 
\begin{equation}
	\begin{aligned}
		\mathcal{L}_{adv} = &-\mathbb{E}\{\log[\mathcal{S}(\,D(Y_{s,t}) - D(X_s)\,)]\} \\
		&- \mathbb{E}\{\log[\mathcal{S}(\,D(X_s) - D(Y_{s,t})\,)]\},
	\end{aligned}
	\label{loss:adv}
\end{equation} 
where $\mathcal{S}(\cdot)$ stands for the sigmoid activation. 

Finally, total objective function with hyperparameters $\{w_1, w_2, w_3, w_4, w_5\}$ is defined as:
\begin{equation}
	\begin{aligned}
		\mathcal{L}_{total} = \mathcal{L}_{adv} &+ w_1\mathcal{L}_{mask} + w_2\mathcal{L}_{kernel} \\
		 &+ w_3\mathcal{L}_{id} + w_4\mathcal{L}_{context}.
	\end{aligned}
	\label{loss:total}	
\end{equation}

Our model, CarTrans, is trained end-to-end on the training set of commonly-used CelebA-HQ~\cite{karras_2018_celebahq} and FlickrFaces-HQ (FFHQ)~\cite{Tero_2019_FFHQ} datasets based on Eq.~\ref{loss:total}.  
For more details on the training strategy and hyperparameter setting please refer to the appendix.

\section{Experiments}

\begin{figure*}[t] 
	\centering
	\vspace{-3mm}
	\includegraphics[width=0.99\linewidth]{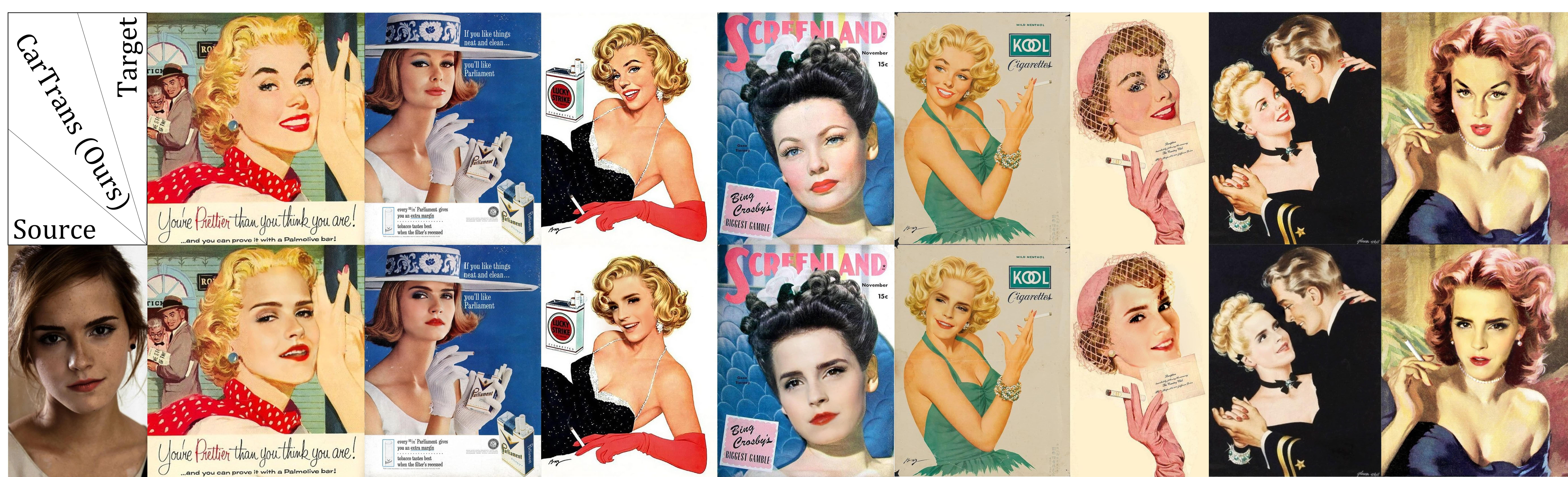}
	\vspace{-2mm}
	\caption{Results of CarTrans, transferring from portrait to vintage advertisements, across large gaps in pose, expression, and appearance.}
	\label{fig:poster}
	\vspace{-3mm}
\end{figure*}

In this section, we first compare our method with \textbf{five SOTA methods}:  FaceSwap~\cite{Korshunova_2017_ICCV}, FSGAN~\cite{Nirkin_2019_ICCV}, Deepfakes~\cite{deepfakes_2020}, FaceShifter~\cite{Li_2020_CVPR}, and InfoSwap~\cite{Gao_2021_CVPR}. 
Experiments are based on \textbf{three datasets}: Face-Forensics++ (FF++)~\cite{roessler_2019_faceforensics}, DeeperForensics-1.0 (DF-v1)~\cite{jiang_2020_deeperforensics}, and the test set of CelebA-HQ and FFHQ. 
We then analyze the role of our proposed CAR and KeRE via an ablation study on FF++ and DF-v1.   
An user study is provided for further evaluation. 

{\bf Quantitative metrics.} We evaluate the performance of each method based on three commonly used quantitative metrics. 
\textbf{Identity Retrieval} stands for the mean accuracy of the generated face in $Y_{s,t}$ being correctly identified as the source individual, 
using a SOTA recognition model~\cite{wang_2018_cosface} (different from $M^{id}(\cdot)$ used in CarTrans for a fair evaluation). 
Also, we show the cosine similarities $\cos\langle z_{Y_{s,t}}, z_s\rangle$ and $\cos\langle z_{Y_{s,t}}, z_t\rangle$ for better demonstration (where $z_{Y_{s,t}} = M^{id}(Y_{s,t})$ for short).  
\textbf{Pose Error} and \textbf{Expression Error} stand for the mean square error (MSE) in the head pose and facial expression of $Y_{s,t}$ corresponding to those of $X_t$, using a SOTA 3D face alignment model~\cite{siggraphAsia_2017_flame} (also different from $M^{3d}(\cdot)$ used in our model).

\subsection{Comparison with SOTA Methods}

\subsubsection{Comparison on FF++}
\begin{figure}[t] 
	\centering
	\includegraphics[width=0.99\linewidth]{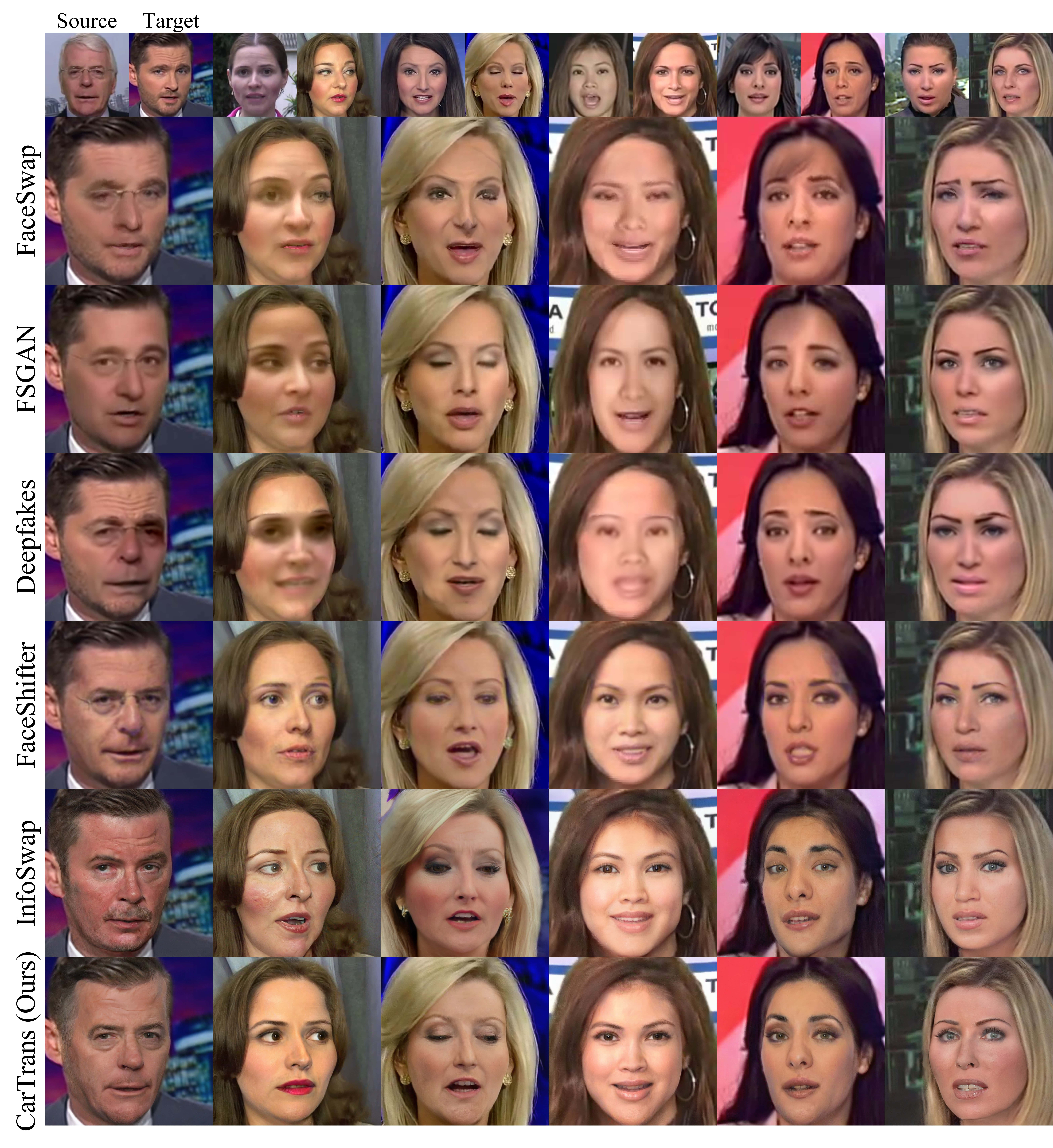}
	\vspace{-3mm}
	\caption{Qualitative comparison with SOTA methods on FF++.}
	\vspace{-4mm}
	\label{fig:ff++}
\end{figure}

The manipulated results of FaceSwap, Deepfakes, and FaceShifter are included in FF++ dataset. Results of FSGAN and InfoSwap are obtained using official pre-trained models. 
Following the same testing protocol with~\cite{Gao_2021_CVPR}, we uniformly extract $60$ frames from each manipulated video, forming an evaluating set of $60$k manipulated frames for each method. The experimental results of quantitative metrics in Tab.~\ref{tab:ff++} indicate that our method outperforms other SOTA methods in both source identity preservation and consistency with target pose and expression. More qualitative examples (at $512\times 512$ resolution) are shown in Fig.~\ref{fig:ff++}. 

\begin{table}[t]  
	\renewcommand{\arraystretch}{1.3}
	\small
	\setlength{\tabcolsep}{2pt}
	\resizebox{\linewidth}{!}{%
		\begin{tabular}{|l|c|cccc|}
			\hline
			\multicolumn{1}{|c|}{\multirow{3}{*}{\textbf{Method}}} & \multicolumn{5}{c|}{\textbf{Quantitative Comparisons on FF++}} 
			\\ \cline{2-6} 
			\multicolumn{1}{|c|}{} & \multicolumn{3}{c|}{Identity Retrieval} & \multicolumn{1}{c|}{\multirow{2}{*}{\makecell{Pose \\ Error $\downarrow$}}} & \multirow{2}{*}{\makecell{Expression\\Error $\downarrow$}} 
			\\ \cline{2-4}
			\multicolumn{1}{|c|}{} & Accuracy $\uparrow$ & $\cos\langle z_{Y_{s,t}}, z_s\rangle$$\uparrow$ & \multicolumn{1}{c|}{$\cos\langle z_{Y_{s,t}}, z_t\rangle$$\downarrow$} & \multicolumn{1}{c|}{} &  
			\\ \hline
			FaceSwap~\cite{Korshunova_2017_ICCV} & {\color{Gray}0.748} & {\color{Gray} 0.436} & \multicolumn{1}{c|}{\color{Gray} 0.319} & \multicolumn{1}{c|}{\color{Gray} 0.0030} & {\color{Gray} 0.050} 
			\\
			FSGAN~\cite{Nirkin_2019_ICCV} & {\color{Gray} 0.635} & \color{Gray} 0.396 & \multicolumn{1}{c|}{\color{Gray}0.358} & \multicolumn{1}{c|}{0.0011} & {0.044} 
			\\
			Deepfakes~\cite{deepfakes_2020} & \color{Gray} 0.866 & \color{Gray}0.487 & \multicolumn{1}{c|}{\color{Gray}0.273} & \multicolumn{1}{c|}{\color{Gray}0.0053} & \color{Gray}0.118 
			\\
			FaceShifter~\cite{Li_2020_CVPR} & \color{Gray} 0.909 & \color{Gray}0.538 & \multicolumn{1}{c|}{\color{Gray}0.317} & \multicolumn{1}{c|}{\color{Gray}0.0029} & \color{Gray}0.055 
			\\
			InfoSwap~\cite{Gao_2021_CVPR} & {0.964} & {0.622} & \multicolumn{1}{c|}{\textbf{0.162}} & \multicolumn{1}{c|}{\color{Gray}0.0031} & \color{Gray}0.048 
			\\ \hline
			\textbf{CarTrans (Ours)} & {\textbf{0.981}} & {\textbf{0.631}} & \multicolumn{1}{c|}{0.213} & \multicolumn{1}{c|}{\textbf{0.0007}} & {\textbf{0.039}} \\ \hline
		\end{tabular}
	}
	\renewcommand{\arraystretch}{1}
	\vspace{-2mm}
	\caption{Comparison with SOTA methods using three quantitative metrics on FF++ dataset. All results are obtained following the same protocol. The best two results are in [\textbf{dark}] and [dark] respectively. $\uparrow$: higher is better. $\downarrow$: lower is the better.}
	\label{tab:ff++}
	\vspace{-2mm}
\end{table}

\begin{table*}[t] 
	\vspace{-3mm}
	\centering
	\renewcommand{\arraystretch}{1.2}
	\small
	\setlength{\tabcolsep}{3pt}
	\resizebox{1.0\linewidth}{!}{%
		\begin{tabular}{|l|c|cccccccccc|c|}
			\hline 
			\multicolumn{1}{|c|}{\multirow{3}{*}{\textbf{Method}}} & \multicolumn{12}{c|}{\textbf{Quantitative Comparisons on DF-v1}} \\ \cline{2-13}			
			\multicolumn{1}{|c|}{} & \multicolumn{3}{c|}{Identity Retrieval} & \multicolumn{1}{c|}{\multirow{2}{*}{\makecell{Pose\\Error $\downarrow$}}} & \multicolumn{8}{c|}{Expression Error $\downarrow$} 
			\\ \cline{2-4}\cline{6-13}  
			\multicolumn{1}{|c|}{} & \multicolumn{1}{c|}{Accuracy $\uparrow$} & $\cos\langle z_{Y_{s,t}}, z_s\rangle$ $\uparrow$ & \multicolumn{1}{c|}{$\cos\langle z_{Y_{s,t}}, z_t\rangle$ $\downarrow$} & \multicolumn{1}{c|}{} & Angry & Contempt & Disgust & Fear & Happy & Sad & Surprise & Overall 
			\\ \cline{1-13} 
			FSGAN~\cite{Nirkin_2019_ICCV} & \multicolumn{1}{c|}{\color{Gray}0.7199} & \color{Gray}0.4289 & \multicolumn{1}{c|}{\color{Gray}0.3051} & \multicolumn{1}{c|}{{0.0019}} & {0.1062} & {0.1026} & {0.1143} & {0.1109} & {0.0899} & {0.0937} & {0.1217} & {0.1035} 
			\\
			FaceShifter~\cite{Li_2020_CVPR} & \multicolumn{1}{c|}{\color{Gray}0.9467} & \color{Gray}0.5349 & \multicolumn{1}{c|}{\color{Gray}0.2973} & \multicolumn{1}{c|}{\color{Gray}0.0024} & \color{Gray}0.1621 & \color{Gray}0.1504 & \color{Gray}0.1677 & \color{Gray}0.1763 & \color{Gray}0.1288 & \color{Gray}0.1364 & \color{Gray}0.1891 & \color{Gray}0.1542 
			\\
			InfoSwap~\cite{Gao_2021_CVPR} & \multicolumn{1}{c|}{0.9734} & {0.7227} & \multicolumn{1}{c|}{\textbf{0.1916}} & \multicolumn{1}{c|}{\color{Gray}0.0022} & \color{Gray}0.1221 & \color{Gray}0.1121 & \color{Gray}0.1195 & \color{Gray}0.1265 & \color{Gray}0.0936 & \color{Gray}0.1141 & \color{Gray}0.1338 & \color{Gray}0.1161 
			\\ \hline
			\textbf{CarTrans (ours)} & \multicolumn{1}{c|}{\textbf{0.9884}} & \textbf{0.7746} & \multicolumn{1}{c|}{0.2034} & \multicolumn{1}{c|}{\textbf{0.0017}} & \textbf{0.0653} & \textbf{0.0616} & \textbf{0.0692} & \textbf{0.0689} & \textbf{0.0617} & \textbf{0.0641} & \textbf{0.0764} & \textbf{0.0652} \\ 
			\hline
		\end{tabular}
	}%
	\renewcommand{\arraystretch}{1}
	\vspace{-2mm}
	\caption{Further comparisons on DF-v1 dataset in terms of matching unobserved $X_t$ with various poses and expressions. Results are obtained following the same protocol. The best two results are in [\textbf{dark}] and [dark] respectively. $\uparrow$: higher is better. $\downarrow$: lower is the better.
	}
	\label{tab:df-v1}
	\vspace{-2mm}
\end{table*}

\begin{figure*}[t] 
	\centering
	\includegraphics[width=0.99\linewidth]{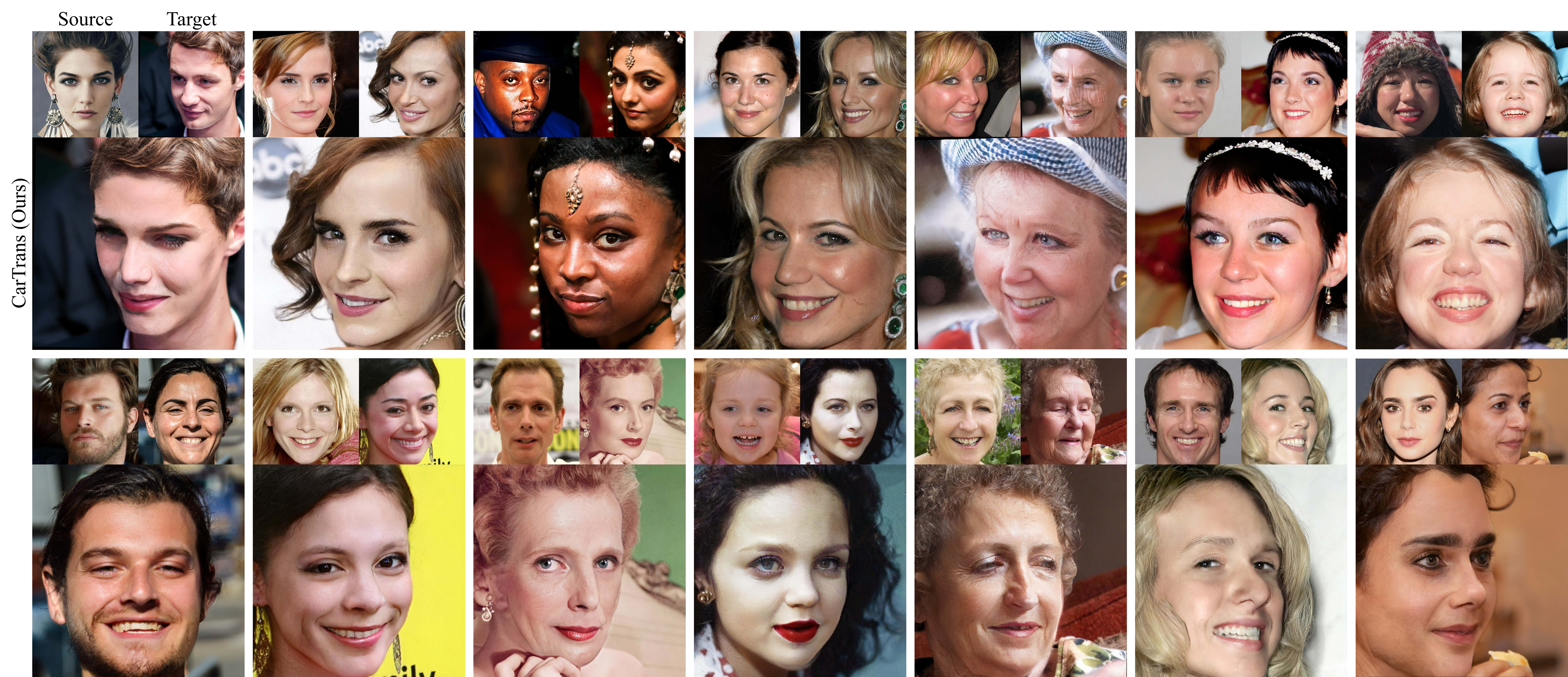}
	\vspace{-2mm}
	\caption{Qualitative results of CarTrans on the test sets of CelebA-HQ and FFHQ, across large gaps in pose and expression. 
	}
	\vspace{-3mm}
	\label{fig:celeb}
\end{figure*}

\vspace{-2mm}
\subsubsection{Comparison on DF-v1}
Since faces in FF++ videos are mainly canonical, we further perform comparisons on DF-v1 dataset which contains videos of $95$ actors performing different poses and expressions. 
Here we mainly compare CarTrans with FSGAN, FaceShifter, and InfoSwap, as they show competitive performance in preserving target expression and pose on FF++. 
Since FaceShifter has not been officially tested on DF-V1 before, and its pre-trained model is currently not released, we use our reproduced version which is trained following exactly the same training protocol as described in~\cite{Li_2020_CVPR} on CelebA-HQ, FFHQ and VGGFace~\cite{parkhi_2015_deep} for $70$ epochs. 
For detailed testing protocol please refer to the appendix. 
We evaluate results of each method using the three quantitative metrics. 
The expression errors are computed both for each expression individually and for an overall performance over all expressions.  
As shown in Tab.~\ref{tab:df-v1}, CarTrans outperforms other three methods in every metric.  
Some qualitative examples of CarTrans (at $512\times 512$ resolution) on wild images and the test set of CelebA-HQ and FFHQ are shown in Fig.~\ref{fig:poster} and Fig.~\ref{fig:celeb}, respectively. 
For qualitative results of CarTrans please refer to the appendix. 

\subsection{User Study on FF++}
\label{subsec:user}

\begin{table}[h]  
	\centering
	\renewcommand{\arraystretch}{1.1}
	\small
	\setlength{\tabcolsep}{3pt}
	\resizebox{\linewidth}{!}{%
		\begin{tabular}{|l|ccc|}
			\hline
			\multicolumn{1}{|c|}{\multirow{2}{*}{\textbf{Method}}} & \multicolumn{3}{c|}{\textbf{User Study on FF++}} \\ \cline{2-4} 
			\multicolumn{1}{|c|}{} & \multicolumn{1}{c|}{Identity $\uparrow$} & \multicolumn{1}{c|}{Pose and Expression $\uparrow$} & Fidelity $\uparrow$ \\ \hline
			FaceSwap~\cite{Korshunova_2017_ICCV} & \multicolumn{1}{c|}{\color{Gray}0.0315} & \multicolumn{1}{c|}{\color{Gray}0.0290} & \color{Gray}0.0240 
			\\
			FSGAN~\cite{Nirkin_2019_ICCV} & \multicolumn{1}{c|}{\color{Gray}0.0170} & \multicolumn{1}{c|}{\color{Gray}0.1305} & \color{Gray}0.0295 
			\\
			Deepfakes~\cite{deepfakes_2020} & \multicolumn{1}{c|}{\color{Gray}0.0465} & \multicolumn{1}{c|}{\color{Gray}0.0270} & \color{Gray}0.0145 
			\\
			FaceShifter~\cite{Li_2020_CVPR} & \multicolumn{1}{c|}{\color{Gray}0.1620} & \multicolumn{1}{c|}{\color{Gray}0.1425} & \color{Gray}0.1370 
			\\
			InfoSwap~\cite{Gao_2021_CVPR} & \multicolumn{1}{c|}{0.3475} & \multicolumn{1}{c|}{0.2345} & {0.3770} 
			\\ \hline
			\textbf{CarTrans (Our)} & \multicolumn{1}{c|}{\textbf{0.3955}} & \multicolumn{1}{c|}{\textbf{0.4365}} & \textbf{0.4180} \\ \hline
		\end{tabular}
	}
	\vspace{-2mm}
	\caption{Comparison via user study. Numbers stand for the frequencies of each method being selected as the best in each aspect. 
	}
	\vspace{-3mm}
	\label{tab:user}
\end{table}

\begin{table*}[t] 
	\vspace{-3mm}
	\centering
	\renewcommand{\arraystretch}{1.2}
	\small
	\setlength{\tabcolsep}{3pt}
	\resizebox{1.0\linewidth}{!}{%
		\begin{tabular}{|l|c|cccccccccc|c|}
			\hline 
			\multicolumn{1}{|c|}{\multirow{3}{*}{\textbf{Configuration}}} & \multicolumn{12}{c|}{\textbf{Ablation Study on DF-v1}} 
			\\ \cline{2-13}			
			\multicolumn{1}{|c|}{} & \multicolumn{3}{c|}{Identity Retrieval} & \multicolumn{1}{c|}{\multirow{2}{*}{\makecell{Pose\\Error $\downarrow$}}} & \multicolumn{8}{c|}{Expression Error $\downarrow$} 
			\\ \cline{2-4}\cline{6-13}  
			\multicolumn{1}{|c|}{} & \multicolumn{1}{c|}{Accuracy $\uparrow$} & $\cos\langle z_{Y_{s,t}}, z_s\rangle$ $\uparrow$ & \multicolumn{1}{c|}{$\cos\langle z_{Y_{s,t}}, z_t\rangle$ $\downarrow$} & \multicolumn{1}{c|}{} & Angry & Contempt & Disgust & Fear & Happy & Sad & Surprise & Overall 
			\\ \cline{1-13} 
			\textbf{CarTrans} & \multicolumn{1}{c|}{\textbf{0.9884}} & \textbf{0.7746} & \multicolumn{1}{c|}{\textbf{0.2034}} & \multicolumn{1}{c|}{\textbf{0.0017}} & \textbf{0.0653} & \textbf{0.0616} & \textbf{0.0692} & \textbf{0.0689} & \textbf{0.0617} & \textbf{0.0641} & \textbf{0.0764} & \textbf{0.0652} 
			\\ \hline
			CarTrans w/o CAR & \multicolumn{1}{c|}{0.9781} & {0.7693} & \multicolumn{1}{c|}{0.2289} & \multicolumn{1}{c|}{\color{Gray}0.0023} & \color{Gray}0.1453 & \color{Gray}0.1109 & \color{Gray}0.1673 & \color{Gray}0.1936 & \color{Gray}0.1707 & \color{Gray}0.1867 & \color{Gray}0.1931 & \color{Gray}0.1565 
			\\
			CarTrans w/o KeRE & \multicolumn{1}{c|}{\color{Gray}0.9143} & \color{Gray}0.6694 & \multicolumn{1}{c|}{\color{Gray}0.2895} & \multicolumn{1}{c|}{0.0018} & 0.0749 & 0.0713 & 0.0788 & 0.0785 & 0.0715 & 0.0738 & 0.0860 & 0.0748 
			\\
			CarTrans w/o (CAR and KeRE) & \multicolumn{1}{c|}{\color{Gray}0.9056} & \color{Gray}0.6618 & \multicolumn{1}{c|}{\color{Gray}0.2928} & \multicolumn{1}{c|}{\color{Gray}0.0024} & \color{Gray}0.1575 & \color{Gray}0.1268 & \color{Gray}0.1645 & \color{Gray}0.1864 & \color{Gray}0.1314 & \color{Gray}0.1416 & \color{Gray}0.2477 & \color{Gray}0.1575 
			\\ \hline
		\end{tabular}
	}%
	\renewcommand{\arraystretch}{1}
	\vspace{-2mm}
	\caption{Comparisons between different configurations of CarTrans on DF-v1 dataset. Results are obtained following the same protocol. The best two results are in [\textbf{dark}] and [dark] respectively. $\uparrow$: higher is better. $\downarrow$: lower is the better.}
	\label{tab:ablation}
	\vspace{-4mm}
\end{table*}

As an additional comparison with SOTA methods, we perform a user study on FF++ dataset, as shown in Tab.~\ref{tab:user}. 
A total of $40$ users are invited. For each user we show $30$ groups of manipulated images, and each group includes $5$ generated results by the $5$ methods using a same random-selected pair of $(X_s, X_t)$.   
Then, from each group,
users are asked to select out $3$ images, \ie: \textcolor{blue}{(a)} the one most similar to $X_s$ in identity ($1^{st}$ column), \textcolor{blue}{(b)} the one most similar to $X_t$ in pose and expression ($2^{nd}$ column), and \textcolor{blue}{(c)} the one with the highest fidelity ($3^{rd}$ column). 
The statistical results of user evaluations also show the superiority of our method.

\subsection{Analysis on Components}

\begin{figure}[H] 
	\centering
	\vspace{-1mm}
	\includegraphics[width=0.99\linewidth]{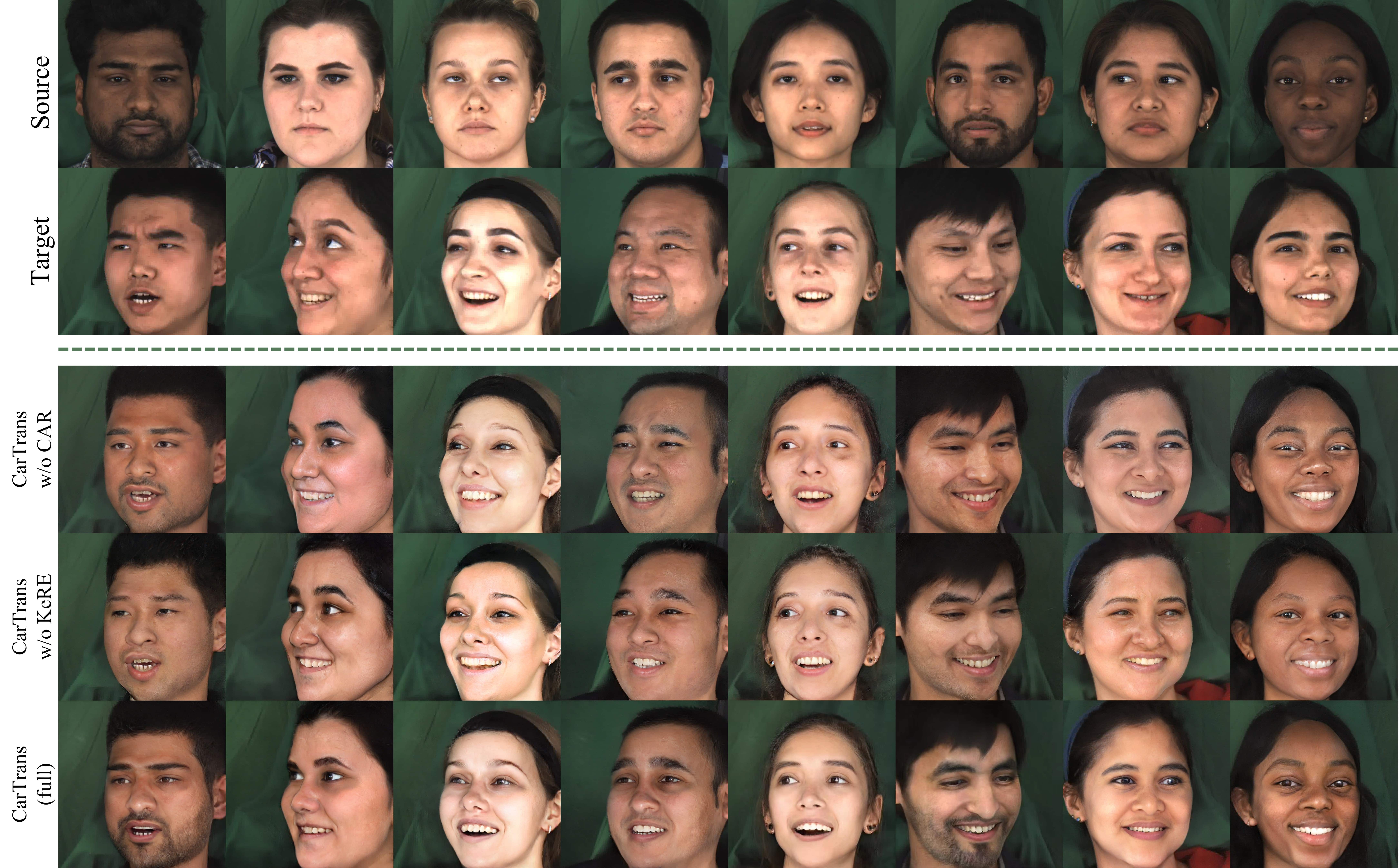}
	\vspace{-1mm}
	\caption{Qualitative examples in the ablation study on DF-v1.}
	\vspace{-2mm}
	\label{fig:df-v1}
\end{figure}

To analyze the role of the proposed CAR and KeRE in improving the performance of CarTrans in synthesizing $Y_{s,t}$ (in Eq.~\ref{eq:generation}), we evaluate three different configurations of CarTrans:  
\textcolor{blue}{(i)} Replacing CAR $z^*_{s,t}$ with original $z_s$ (CAR w/o CAR), 
\textcolor{blue}{(ii)} replacing identity-agnostic context representations $\{H_t^{(i)}\}$ with original $\{F_t^{(i)}\}$ (CarTrans w/o KeRE), 
and \textcolor{blue}{(iii)} CarTrans w/o (CAR and KeRE). 
These three configurations are trained following the same protocol as the full model CarTrans. 
Some qualitative examples are shown in Fig.~\ref{fig:df-v1}.
As shown in Tab.~\ref{tab:ablation}, all quantitative metrics are down compared to the full model. 
Among them, \textit{identity retrieval} drops off significantly when $\{H_t^{(i)}\}$ are replace by $\{F_t^{(i)}\}$ (CarTrans w/o KeRE), 
while \textit{pose and expression errors} rise more when the context-aware corrective are removed from face representation (CarTrans w/o CAR). 
These results suggest that both CAR and KeRE play an important role in improving performance of CarTrans, while each has a different impact. 

%

\section{Related work}
Recent advances in controllable face image synthesis have yielded many impressive applications~\cite{thies_2016_face2face, bao2018towards, thies_2019_deferred, deng_2020_disentangled, Buhler_2021_ICCV, Xia_2021_CVPR, zhizhong_2021_CVPR, Shi_2021_CVPR, Yue_2021_CVPR, Li_2021_CVPR}.
In the subarea of face transfer, research begins with the influential work~\cite{blanz_2004_exchanging} which focuses on cases where the viewpoint and illumination of target images are different from the source. 
FaceSwap~\cite{Korshunova_2017_ICCV} proposes a case-by-case model to enable real-time face transfer. 
\cite{Nirkin_2018_IEEE} highlights the model robustness under unpredictable conditions. 

More recently, 2D GAN-based methods have dominated this field for their superior capacity of synthesis photorealistic images.  
For instance, FSGAN~\cite{Nirkin_2019_ICCV} achieves subject-agnostic identity transfer and considers target skin color and lighting condition preservation. 
FaceShifter~\cite{Li_2020_CVPR} focuses on preserving the target facial occlusions based on a second-stage learning. 
AOT~\cite{zhu_2020_aot} proposes a post-processing method to reduce the discrepancies in lighting and skin color. 
InfoSwap~\cite{Gao_2021_CVPR} contributes a disentanglement method for generating more identity-discriminative results.
At the same time, a number of important advances have been made in face forgery detection~\cite{nguyen_2019_multi, roessler_2019_faceforensics, jiang_2020_deeperforensics, li_2020_celeb, li_2020_facex} to prevent the misuse of face synthesis methods.  

So far, in controllable face synthesis, it is yet a challenge to correctly adapt the identity-dependent face shape and appearance of an individual to various poses and expressions. 
To the best of our knowledge, CarTrans is the first method that learns a context-aware face representation, which allows to causally reason about the potential changes in face appearance in responds to different target contexts. 

\section{Broad Impact}
One of the most fundamental components of computer vision is image synthesis. Recently, synthesizing photorealistic images is no longer sufficient for new applications of generative models, but the properties of the images need also be controllable during synthesis~\cite{Liao2020CVPR, giraffe2021}. 
A specific focus lies on face transfer, a cutting-edge technology in synthetic media that edits the appearance of a person in an existing face image or video according to specified requirements. This topic is a currently vibrant sub-field of controllable image synthesis, due to its application value in video games and film production.
However, if misused, this technology can also raise serious privacy or public issue. 
We further discus broader impacts of face transfer and ways to reduce potential negative social impacts in the appendix.

%
%
%

\section{Conclusion}
In this work, we have introduced CarTrans, a novel generative model of face transfer. 
We have shown that by explicitly incorporating a context-aware corrective into face representation based on prior knowledge about the target context, 
and eliminating the identity specificity of target face via a set of kernel-based regressor when encoding target contextual information, 
CarTrans is capable of adapting face representations according to different target images, 
enabling fine-grained synthesis of identity-dependent face shape and appearance 
across large gaps in the context with high fidelity.  
The experimental results provide empirical evidence that, by characterizing the causal variables in unobserved novel contexts, our proposed context-aware face representation is better suited for generative modeling of human faces in the context of face transfer. 

{\small
\bibliographystyle{ieee_fullname}
\bibliography{egbib}

\begin{thebibliography}{10}\itemsep=-1pt

\bibitem{deepfakes_2020}
{\em Deepfakes}, Accessed: 2021-10-03.
\newblock \url{https://github.com/deepfakes/faceswap}.

\bibitem{bao2018towards}
Jianmin Bao, Dong Chen, Fang Wen, Houqiang Li, and Gang Hua.
\newblock Towards open-set identity preserving face synthesis.
\newblock In {\em CVPR}, 2018.

\bibitem{blanz_2004_exchanging}
Volker Blanz, Kristina Scherbaum, Thomas Vetter, and Hans-Peter Seidel.
\newblock Exchanging faces in images.
\newblock {\em Computer Graphics Forum}, 2004.

\bibitem{Buhler_2021_ICCV}
Marcel~C. B\"uhler, Abhimitra Meka, Gengyan Li, Thabo Beeler, and Otmar
  Hilliges.
\newblock Varitex: Variational neural face textures.
\newblock In {\em ICCV}, 2021.

\bibitem{davidson_2018_hyperspherical}
Tim~R Davidson, Luca Falorsi, Nicola De~Cao, Thomas Kipf, and Jakub~M Tomczak.
\newblock Hyperspherical variational auto-encoders.
\newblock {\em arXiv preprint arXiv:1804.00891}, 2018.

\bibitem{Deng_2019_CVPR}
Jiankang Deng, Jia Guo, Niannan Xue, and Stefanos Zafeiriou.
\newblock Arcface: Additive angular margin loss for deep face recognition.
\newblock In {\em CVPR}, 2019.

\bibitem{deng_2020_disentangled}
Yu Deng, Jiaolong Yang, Dong Chen, Fang Wen, and Xin Tong.
\newblock Disentangled and controllable face image generation via 3d
  imitative-contrastive learning.
\newblock In {\em CVPR}, 2020.

\bibitem{gafni_2019_live}
Oran Gafni, Lior Wolf, and Yaniv Taigman.
\newblock Live face de-identification in video.
\newblock In {\em ICCV}, 2019.

\bibitem{Gao_2021_CVPR}
Gege Gao, Huaibo Huang, Chaoyou Fu, Zhaoyang Li, and Ran He.
\newblock Information bottleneck disentanglement for identity swapping.
\newblock In {\em CVP)}, 2021.

\bibitem{Yue_2021_CVPR}
Yue Gao, Fangyun Wei, Jianmin Bao, Shuyang Gu, Dong Chen, Fang Wen, and Zhouhui
  Lian.
\newblock High-fidelity and arbitrary face editing.
\newblock In {\em CVPR}, 2021.

\bibitem{bfm_2018}
Thomas Gerig, Andreas Morel-Forster, Clemens Blumer, Bernhard Egger, Marcel
  Luthi, Sandro Schoenborn, and Thomas Vetter.
\newblock Morphable face models - an open framework.
\newblock In {\em FG 2018}, 2018.

\bibitem{guo_2020_3ddfa}
Jian-Zhu Guo, Xiang-Yu Zhu, Yang Yang, Fan Yang, Zhen Lei, and Stan~Z Li.
\newblock Towards fast, accurate and stable 3d dense face alignment.
\newblock In {\em ECCV}, 2020.

\bibitem{holland_1986_statistics}
Paul~W Holland.
\newblock Statistics and causal inference.
\newblock {\em JASA}, 1986.

\bibitem{Huang_2017_ICCV}
Xun Huang and Serge Belongie.
\newblock Arbitrary style transfer in real-time with adaptive instance
  normalization.
\newblock In {\em ICCV}, 2017.

\bibitem{zhizhong_2021_CVPR}
Zhizhong Huang, Junping Zhang, and Hongming Shan.
\newblock When age-invariant face recognition meets face age synthesis: A
  multi-task learning framework.
\newblock In {\em CVPR}, 2021.

\bibitem{jiang_2020_deeperforensics}
Liming Jiang, Ren Li, Wayne Wu, Chen Qian, and Chen~Change Loy.
\newblock Deeperforensics-1.0: A large-scale dataset for real-world face
  forgery detection.
\newblock In {\em CVPR}, 2020.

\bibitem{Alexia_2018_RSGAN}
Alexia Jolicoeur-Martineau.
\newblock The relativistic discriminator: a key element missing from standard
  {GAN}.
\newblock In {\em ICLR}, 2019.

\bibitem{karras_2018_celebahq}
Tero Karras, Timo Aila, Samuli Laine, and Jaakko Lehtinen.
\newblock Progressive growing of {GAN}s for improved quality, stability, and
  variation.
\newblock In {\em ICLR}, 2018.

\bibitem{Tero_2019_FFHQ}
Tero Karras, Samuli Laine, and Timo Aila.
\newblock A style-based generator architecture for generative adversarial
  networks.
\newblock In {\em CVPR}, 2019.

\bibitem{karras_2020_analyzing}
Tero Karras, Samuli Laine, Miika Aittala, Janne Hellsten, Jaakko Lehtinen, and
  Timo Aila.
\newblock Analyzing and improving the image quality of stylegan.
\newblock In {\em CVPR}, 2020.

\bibitem{klambauer_2017_self}
G{\"u}nter Klambauer, Thomas Unterthiner, Andreas Mayr, and Sepp Hochreiter.
\newblock Self-normalizing neural networks.
\newblock In {\em NeurIPS}, 2017.

\bibitem{Korshunova_2017_ICCV}
Iryna Korshunova, Wenzhe Shi, Joni Dambre, and Lucas Theis.
\newblock Fast face-swap using convolutional neural networks.
\newblock In {\em ICCV}, 2017.

\bibitem{Li_2021_CVPR}
Jia Li, Zhaoyang Li, Jie Cao, Xingguang Song, and Ran He.
\newblock Faceinpainter: High fidelity face adaptation to heterogeneous
  domains.
\newblock In {\em CVPR}, 2021.

\bibitem{Li_2020_CVPR}
Lingzhi Li, Jianmin Bao, Hao Yang, Dong Chen, and Fang Wen.
\newblock Advancing high fidelity identity swapping for forgery detection.
\newblock In {\em CVPR}, 2020.

\bibitem{li_2020_facex}
Ling-Zhi Li, Jian-Min Bao, Ting Zhang, Hao Yang, Dong Chen, Fang Wen, and
  Bai-Ning Guo.
\newblock Face x-ray for more general face forgery detection.
\newblock In {\em CVPR}, 2020.

\bibitem{siggraphAsia_2017_flame}
Tianye Li, Timo Bolkart, Michael.~J. Black, Hao Li, and Javier Romero.
\newblock Learning a model of facial shape and expression from {4D} scans.
\newblock {\em TOG}, 2017.

\bibitem{li_2020_celeb}
Yue-Zun Li, Xin Yang, Pu Sun, Hong-Gang Qi, and Si-Wei Lyu.
\newblock Celeb-df: A large-scale challenging dataset for deepfake forensics.
\newblock In {\em CVPR}, 2020.

\bibitem{Liao2020CVPR}
Yiyi Liao, Katja Schwarz, Lars Mescheder, and Andreas Geiger.
\newblock Towards unsupervised learning of generative models for 3d
  controllable image synthesis.
\newblock In {\em CVPR}, 2020.

\bibitem{natsume_2018_rsgan}
Ryota Natsume, Tatsuya Yatagawa, and Shigeo Morishima.
\newblock {RSGAN}: Face swapping and editing using face and hair representation
  in latent spaces.
\newblock In {\em SIGGRAPH}, 2018.

\bibitem{nguyen_2019_multi}
Huy~H. Nguyen, Fuming Fang, Junichi Yamagishi, and Isao Echizen.
\newblock Multi-task learning for detecting and segmenting manipulated facial
  images and videos.
\newblock In {\em BTAS}, 2019.

\bibitem{giraffe2021}
Michael Niemeyer and Andreas Geiger.
\newblock {GIRAFFE}: Representing scenes as compositional generative neural
  feature fields.
\newblock In {\em CVPR}, 2021.

\bibitem{Nirkin_2019_ICCV}
Yuval Nirkin, Yosi Keller, and Tal Hassner.
\newblock Fsgan: Subject agnostic face swapping and reenactment.
\newblock In {\em ICCV}, 2019.

\bibitem{Nirkin_2018_IEEE}
Yuval Nirkin, Iacopo Masi, Anh~Tran Tuan, Tal Hassner, and Gerard Medioni.
\newblock On face segmentation, face swapping, and face perception.
\newblock In {\em FG}, 2018.

\bibitem{Park_2019_CVPR}
Taesung Park, Ming-Yu Liu, Ting-Chun Wang, and Jun-Yan Zhu.
\newblock Semantic image synthesis with spatially-adaptive normalization.
\newblock In {\em CVPR}, 2019.

\bibitem{parkhi_2015_deep}
Omkar~M Parkhi, Andrea Vedaldi, and Andrew Zisserman.
\newblock Deep face recognition.
\newblock In {\em BMVC}, 2015.

\bibitem{pearl_2018_book}
Judea Pearl and Dana Mackenzie.
\newblock {\em The book of why: the new science of cause and effect}.
\newblock Basic Books, 2018.

\bibitem{roessler_2019_faceforensics}
Andreas R\"ossler, Davide Cozzolino, Luisa Verdoliva, Christian Riess, Justus
  Thies, and Matthias Nie{\ss}ner.
\newblock Face{F}orensics++: Learning to detect manipulated facial images.
\newblock In {\em ICCV}, 2019.

\bibitem{rubin_1974_estimating}
Donald~B Rubin.
\newblock Estimating causal effects of treatments in randomized and
  nonrandomized studies.
\newblock {\em Journal of Educational Psychology (APA)}, 1974.

\bibitem{rubin_2005_causal}
Donald~B Rubin.
\newblock Causal inference using potential outcomes: Design, modeling,
  decisions.
\newblock {\em JASA}, 2005.

\bibitem{Schulz_2020_Restricting}
Karl Schulz, Leon Sixt, Federico Tombari, and Tim Landgraf.
\newblock Restricting the flow: Information bottlenecks for attribution.
\newblock In {\em ICLR}, 2020.

\bibitem{Shi_2021_CVPR}
Yichun Shi, Divyansh Aggarwal, and Anil~K. Jain.
\newblock Lifting 2d stylegan for 3d-aware face generation.
\newblock In {\em CVPR}, 2021.

\bibitem{Ravid_2017_opening}
Ravid Shwartz-Ziv and Naftali Tishby.
\newblock Opening the black box of deep neural networks via information.
\newblock {\em arXiv preprint arXiv:1703.00810}, 2017.

\bibitem{thies_2019_deferred}
Justus Thies, Michael Zollh{\"o}fer, and Matthias Nie{\ss}ner.
\newblock Deferred neural rendering: Image synthesis using neural textures.
\newblock {\em TOG}, 2019.

\bibitem{thies_2016_face2face}
Justus Thies, Michael Zollh{\"o}fer, Marc Stamminger, Christian Theobalt, and
  Matthias Nie{\ss}ner.
\newblock Face2face: Real-time face capture and reenactment of rgb videos.
\newblock In {\em CVPR}, 2016.

\bibitem{Tishby_2015_IB}
Naftali Tishby and Noga Zaslavsky.
\newblock Deep learning and the information bottleneck principle.
\newblock In {\em ITW}, 2015.

\bibitem{Oord_2017_vqvae}
Aaron van~den Oord, Oriol Vinyals, and koray kavukcuoglu.
\newblock Neural discrete representation learning.
\newblock In {\em NeurIPS}, 2017.

\bibitem{wang_2018_cosface}
Hao Wang, Yi-Tong Wang, Zheng Zhou, Xing Ji, Di-Hong Gong, Jing-Chao Zhou,
  Zhi-Feng Li, and Wei Liu.
\newblock Cosface: Large margin cosine loss for deep face recognition.
\newblock In {\em CVPR}, 2018.

\bibitem{Xia_2021_CVPR}
Weihao Xia, Yujiu Yang, Jing-Hao Xue, and Baoyuan Wu.
\newblock Tedigan: Text-guided diverse face image generation and manipulation.
\newblock In {\em CVPR}, 2021.

\bibitem{xu_2018_spherical}
Jiacheng Xu and Greg Durrett.
\newblock Spherical latent spaces for stable variational autoencoders.
\newblock {\em arXiv preprint arXiv:1808.10805}, 2018.

\bibitem{zhu_2020_aot}
Hao Zhu, Chaoyou Fu, Qianyi Wu, Wayne Wu, Chen Qian, and Ran He.
\newblock {AOT}: Appearance optimal transport based identity swapping for
  forgery detection.
\newblock {\em NeurIPS}, 2020.

\end{thebibliography}
}

\end{document}